\documentclass[10pt,twocolumn,letterpaper,llncs]{article}

\usepackage{iccv}
\usepackage{times}
\usepackage{epsfig}
\usepackage{graphicx}
\usepackage{amsmath}
\usepackage{amssymb}
\usepackage{bm}
\usepackage{authblk}

\makeatletter
\newcommand{\printfnsymbol}[1]{%
	\textsuperscript{\@fnsymbol{#1}}%
}
\makeatother

\usepackage[pagebackref=true,breaklinks=true,letterpaper=true,colorlinks,bookmarks=false]{hyperref}

 \iccvfinalcopy 

\def\iccvPaperID{6125} 

\ificcvfinal\fi
\begin{document}

\title{Landmark Assisted CycleGAN for Cartoon Face Generation}

\author[1]{\thanks{Equal contribution}Ruizheng Wu}
\author[2]{\printfnsymbol{1}Xiaodong Gu}
\author[3]{Xin Tao}
\author[3]{Xiaoyong Shen}
\author[3]{Yu-Wing Tai}
\author[1,3]{Jiaya Jia}

\affil[1]{The Chinese University of Hong Kong}
\affil[2]{Harbin Institute of Technology, Shenzhen}
\affil[3]{YouTu Lab, Tencent}

\affil[ ]{\tt\small {\{rzwu, leojia\}@cse.cuhk.edu.hk, \{xintao, dylanshen, yuwingtai\}@tencent.com}, guxiaodong@stu.hit.edu.cn}

\maketitle

\begin{abstract}
	In this paper, we are interested in generating an cartoon face of a person by using unpaired training data between real faces and cartoon ones. A major challenge of this task is that the structures of real and cartoon faces are in two different domains, whose appearance differs greatly from each other. Without explicit correspondence, it is difficult to generate a high quality cartoon face that captures the essential facial features of a person. In order to solve this problem, we propose landmark assisted CycleGAN, which utilizes face landmarks to define landmark consistency loss and to guide the training of local discriminator in CycleGAN. To enforce structural consistency in landmarks, we utilize the conditional generator and discriminator. Our approach is capable to generate high-quality cartoon faces even indistinguishable from those drawn by artists and largely improves state-of-the-art.
\end{abstract}

\vspace{-0.2in}
\section{Introduction}
\vspace{-0.1in}
Cartoon faces appear in animations, comics and games. They are widely used as profile pictures in social media platforms, such as Facebook and Instagram. Drawing an cartoon face is labor intensive. Not only it requires professional skills, but also it is difficult to resemble unique appearance of each person. In this paper, we aim at generating alike cartoon faces for any persons automatically. We cast this problem as an image-to-image translation task. However, we consider unpaired training data between cartoon and real faces.

Image-to-image translation was first introduced by Isola et al.~\cite{isola2017image}, which utilizes the generative adversarial network (GAN) \cite{goodfellow2014generative} to translate an image from a source domain to a target domain such that the translated images are close to the ground truth measured by a discriminator network. This method and the follow-up works~\cite{wang2018high,karacan2016learning,sangkloy2017scribbler} require paired data for training. However, it is not always easy to obtain a large amount of paired data. Thus, CycleGAN~\cite{zhu2017unpaired} was introduced. It uses the cycle consistency loss to train two pairs of generators and discriminators in order to regularize the solution of trained networks. CycleGAN demonstrated impressive results, such as ``horse-to-zebra'' conversion.

\begin{figure}[t!]
	\centering
	\begin{tabular}
		{@{\hspace{0.0mm}}c@{\hspace{3mm}}c@{\hspace{2mm}}c@{\hspace{0.0mm}}}
		\includegraphics[width=0.25\linewidth]{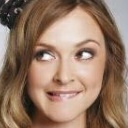}&
		\includegraphics[width=0.25\linewidth]{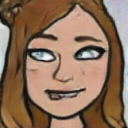}&
		\includegraphics[width=0.25\linewidth]{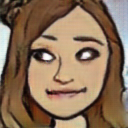}\\
		
		\includegraphics[width=0.25\linewidth]{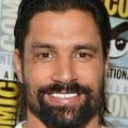}&
		\includegraphics[width=0.25\linewidth]{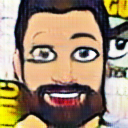}&
		\includegraphics[width=0.25\linewidth]{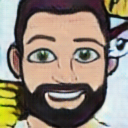}\\
		
		Input&CycleGAN\cite{zhu2017unpaired}&Ours\\
		
	\end{tabular}
	\caption{Given a real-face image, our goal is to generate its corresponding cartoon face, which preserves necessary attributes. Our landmark assisted CycleGAN generates visually plausible results. Note the similarities of hair style, face shape, eyes and mouths of our generated cartoon faces in comparisons with the real faces.}
	\label{case_fig}
\end{figure}

In our ``face-to-cartoon'' conversion, we found directly applying CycleGAN cannot produce satisfactory results, as shown in Fig. \ref{case_fig}. This is because the geometric structures of the two domains are so different from each other, which make mismatching of structures leading to severe distortion and visual artifacts. To address the geometric inconsistency problem, we propose to incorporate more spatial structure information into current framework. More specifically, landmark information is the effective sparse spatial constraint which mitigate this problem, and multiple strategies can be adopted with it to tackle the geometric issues.

We thus propose landmark assisted CycleGAN where face landmarks of real and cartoon faces are used in conjunction with the original images of real and cartoon faces. We design a landmark consistency loss and landmark matched global discriminator to enforce the similarity of facial structures. The explicit structural constraints in the two domains ensure that semantic properties, e.g. eyes, nose, and mouth, can still be matched correctly even without paired training data. This effectively avoids distortion of facial structures in the generated cartoon images. 
In addition, face landmarks can be used to define local discriminators, which further guide the training of generators to pay more attention to important facial features for visually more plausible result generation. The main contributions of our work is multifold.
\vspace{-0.1in}
\begin{itemize}
	\item We propose a landmark assisted CycleGAN to translate real faces into cartoon faces with unpaired training data. It produces significantly higher-quality results than the original CycleGAN with our understanding of this special problem and corresponding system design.
	
	\item We introduce the landmark consistency loss, which effectively solve the problem of structural mismatching between unpaired training data.
	
	\item We use global and local discriminators that notably enhance the quality of generated cartoon faces.
	
	\item We build a new dataset with two kinds of cartoon styles. This dataset contains 2,125 samples for bitmoji styles and 17,920 images for anime faces style respectively, and corresponding landmarks are annotated for both two styles.
	
\end{itemize} 

\section{Related work}
\vspace{-0.05in}
\subsection{Generative Adversarial Networks}
\vspace{-0.05in}
Generative adversarial networks (GANs) \cite{goodfellow2014generative,arjovsky2017wasserstein,radford2015unsupervised} have produced impressive results in many computer vision tasks, such as image generation \cite{denton2015deep,radford2015unsupervised}, super-resolution \cite{ledig2017photo}, image editing \cite{zhu2016generative}, image synthesis \cite{wang2018high} and several other tasks. A GAN contains generator and discriminator networks. They are trained with adversarial loss, which forces the generated images to be similar to real images in the training data. 

In order to place more control to the generation process, variations of GANs were proposed, such as CGAN \cite{mirza2014conditional,miyato2018cgans} and ACGAN \cite{odena2017conditional}. They generally take extra information (such as labels and attributes) as part of the input to satisfy specific conditions. In our work, we also apply adversarial loss to constrain generation. We make the conditions applied both globally and locally to much improve the effectiveness of the solution. 


\subsection{Image-to-Image Translation}
\vspace{-0.05in}
Image-to-Image translation aims to transform an image from the source domain to target. It involves paired- and unpaired-data translation. For paired data, pix2pix~\cite{isola2017image} applies adversarial loss with L1-loss to train the network.

For unpaired data, there is no corresponding ground truth in target domain. Thus it is more difficult. CoGAN \cite{liu2016coupled} learned a common representation of two domains by sharing weights in generators and discriminators. UNIT \cite{liu2017unsupervised} extended the framework of CoGAN by combining variational auto-encoder (VAE) \cite{kingma2013auto} with adversarial networks. This approach has a strong assumption that different domains should share the same low-dimensional representation in the network. XGAN \cite{royer2017xgan} shares similar structure with UNIT~\cite{liu2017unsupervised} and it introduced the semantic consistency component in feature-level contrary to previous work of using pixel-level consistency. With a single auto-encoder to learn a common representation of different domains, DTN \cite{taigman2016unsupervised,wolf2017unsupervised} transformed images in domains. But it needs a well-pretrained encoder. To overcome the limitations of above methods, more frameworks \cite{murez2018image, li2018unsupervised, gokaslan2018improving, huang2018multimodal, lee2018diverse, mechrez2018contextual} are proposed to improve the results generated by above frameworks.

There are also other methods~\cite{hertzmann2001image,gatys2016image,johnson2016perceptual,gatys2016preserving,ulyanov2016texture,liao2017visual} for image translation. Neural style transfer \cite{gatys2016image,johnson2016perceptual,gatys2016preserving,ulyanov2016texture} synthesized an image with texture of one image and content of another. Deep-Image-Analogy \cite{liao2017visual} is a patch-match \cite{barnes2009patchmatch} based method on high-level feature space and achieved good results in many cases. We note Deep-Image-Analogy is still vulnerable to great variance between the two domains since in this case it is not easy to find high-quality correspondence with patch-match.

\section{Our Method}
\vspace{-0.05in}
\begin{figure*}[t!]
	\centering
	\includegraphics[scale=0.54]{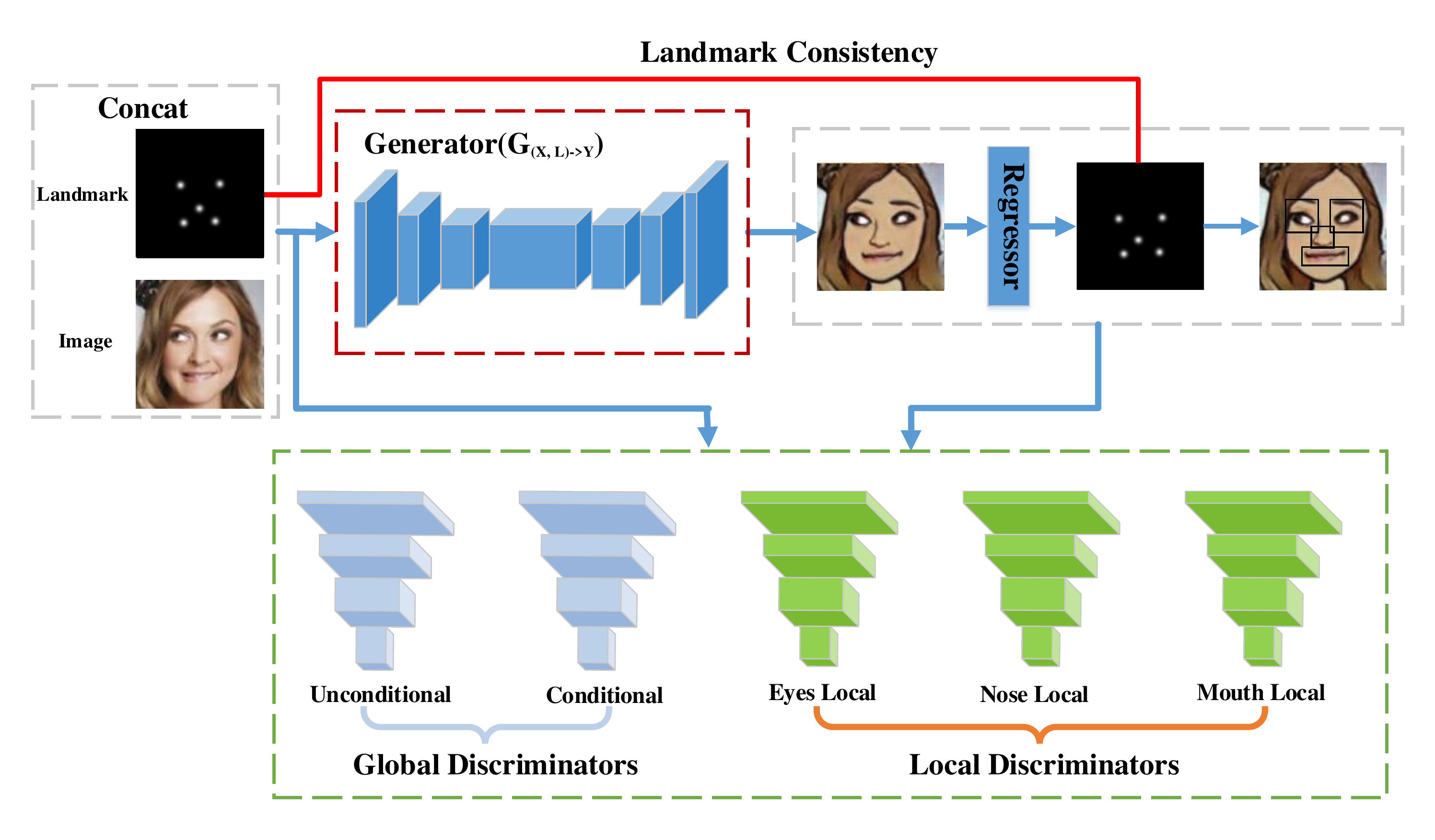}
	\caption{Architecture of our cartoon-face landmark-assisted CycleGAN. Here we only demonstrate the translation part from human to cartoon faces while the counter part from cartoon to human faces is similar. First, the generator outputs coarse cartoon faces. Then a pre-trained regressor predicts facial landmarks. We enforce landmark consistency and local discriminator to solve the problem that huge difference exists regarding structure of two domains. It finally produces realistic and user-specific cartoon faces.}
	\vspace{-0.15in}
	\label{fig:framework}
\end{figure*}

\subsection{Review of CycleGAN}
\vspace{-0.05in}
CycleGAN \cite{zhu2017unpaired} is the base model of our framework, which contains similar structure with DualGAN \cite{yi2017dualgan}. It learns a mapping between domains $X$ and $Y$ given unpaired training samples $x \in X $ and $y \in Y$. For the mapping $G_{X \rightarrow Y}$ and its discriminator $D_Y$, the adversarial loss is defined as
\begin{equation}
\begin{split}
{\mathcal{L}_{GAN}}(G_{X \rightarrow Y}, D_{Y}) = 
 \mathbb{E}_{y} [\log D_{Y}(y)]  \\
~~~ + \mathbb{E}_{x}[\log (1-D_{Y}(G_{X \rightarrow Y}(x)))]
\end{split}
\end{equation}
Different from common GANs, CycleGAN learns the forward and backward mapping simultaneously. Learning of the two mappings are connected by the cycle consistency loss, and the objective function is defined as
\begin{equation}\label{eq_cyc_consistency}
\begin{split}
\mathcal{L}_{cyc} 
= &
||G_{Y \rightarrow X}(G_{X \rightarrow Y}(x))-x||_1 + \\
&||G_{X \rightarrow Y}(G_{Y \rightarrow X}(y))-y||_1.
\end{split}
\end{equation}
The total objective function of CycleGAN is
\begin{equation}\label{eq:cyclegan}
\begin{split}
\mathcal{L}(G_{X \rightarrow Y}, &G_{Y \rightarrow X}, D_{X}, D_{Y}) =  \\
&\mathcal{L}_{GAN}(G_{X \rightarrow Y}, D_{Y}) + \\
&\mathcal{L}_{GAN}(G_{Y \rightarrow X}, D_{X}) + \lambda\mathcal{L}_{cyc}.
\end{split}
\end{equation} 
With the additional cycle consistency loss, CycleGAN achieves impressive results on image translation. However, on our task, it does not perform similarly well since there is a great structural disagreement between source and target domains as afore explained and demonstrated.

In this paper, following Equation (\ref{eq:cyclegan}) where $X$ and $Y$ denote the real- and cartoon-face domains respectively, we first introduce the new landmark assisted cycleGAN (Sec. \ref{sec:assisted}), which is consist of three main parts for enforcing landmark consistency, landmark as condition and landmark guided discriminators. Then we describe our specific training strategies (Sec. \ref{sec:training}). An overview of our framework is shown in Fig. \ref{fig:framework}.


\subsection{Cartoon Face Landmark Assisted CycleGAN}\label{sec:assisted}
\vspace{-0.05in}
Our Landmark assisted part consists of three components, namely landmark consistency loss, landmark-matched global discriminator, and landmark-guided local discriminator.
\vspace{-0.15in}
\subsubsection{Landmark Consistency Loss}
\vspace{-0.05in}
We first give constraints on the real landmark and predicted landmark. We use $\mathcal{L}_2$ norm to compute the loss $\mathcal{L}_{c}$ as:
\begin{equation}\label{eq:lm_consistency}
\begin{split}
{\mathcal{L}_{c}}(&G_{(X,L)\rightarrow Y}) = \\
& \left\| R_{Y}(G_{(X,L) \rightarrow Y}(x, l)) - l  \right\|_2.
\end{split}
\end{equation}
Where $\bm{L}$ indicates the input landmark heatmap set ($l \in \bm{L}$) and $R$ refers to a pre-trained U-Net like landmark regressor with 5-channel output for respective domain, while $R_{Y}$  are used for domain $Y$. 




With the constraint of Equation (\ref{eq:lm_consistency}), we make the images in different domains present close facial structures. Besides, we introduce explicit correspondence between real and cartoon faces with Equation (\ref{eq:lm_consistency}).

\vspace{-0.15in}
\subsubsection{Landmark Matched Global Discriminator}
\vspace{-0.05in}
As shown in Fig. \ref{fig:framework}, we have two global discriminators, which focus differently. For the translation of $X \rightarrow Y$, unconditional global discriminator $\bm{D_{Y}}$ produces more realistic cartoon faces, while conditional global discriminator $\bm{D_{Y}^{g_c}}$ aims to generate landmark-matched cartoon faces with landmark heat map $ l \in  \bm{L}$ as part of input. The objective function of conditional discriminator is
\begin{equation}\label{eq:global_c_d}
\begin{split}
{\mathcal{L}_{GAN}}&(G_{(X,L) \rightarrow Y}, D_{Y}^{g_c}) = \mathbb{E}_{y} [\log D_{Y}(y,l)]  \\
& + \mathbb{E}_{x}[\log (1-D_{Y}(G_{(X,L) \rightarrow Y}(x,l),l))].
\end{split}
\end{equation}
Considering the special design of conditional discriminator, we can adopt a better training strategy on fake sample collection.
Specifically, for ${D_{Y}^{g_c}}$, 
we add cartoon faces with corresponding \emph{unmatched} landmark heat map as additional fake samples to force the generator to produce better matched cartoon faces, otherwise discriminator may considers the landmark-unmatched pairs are also real samples. 
We produce the unmatched pairs by randomly cropping cartoon images to change the position of facial structure, and yet keeping the original landmark coordinates.


\subsubsection{Landmark Guided Local Discriminator}
\vspace{-0.05in}
\begin{figure}[!t]
	\centering
	\includegraphics[scale=0.3]{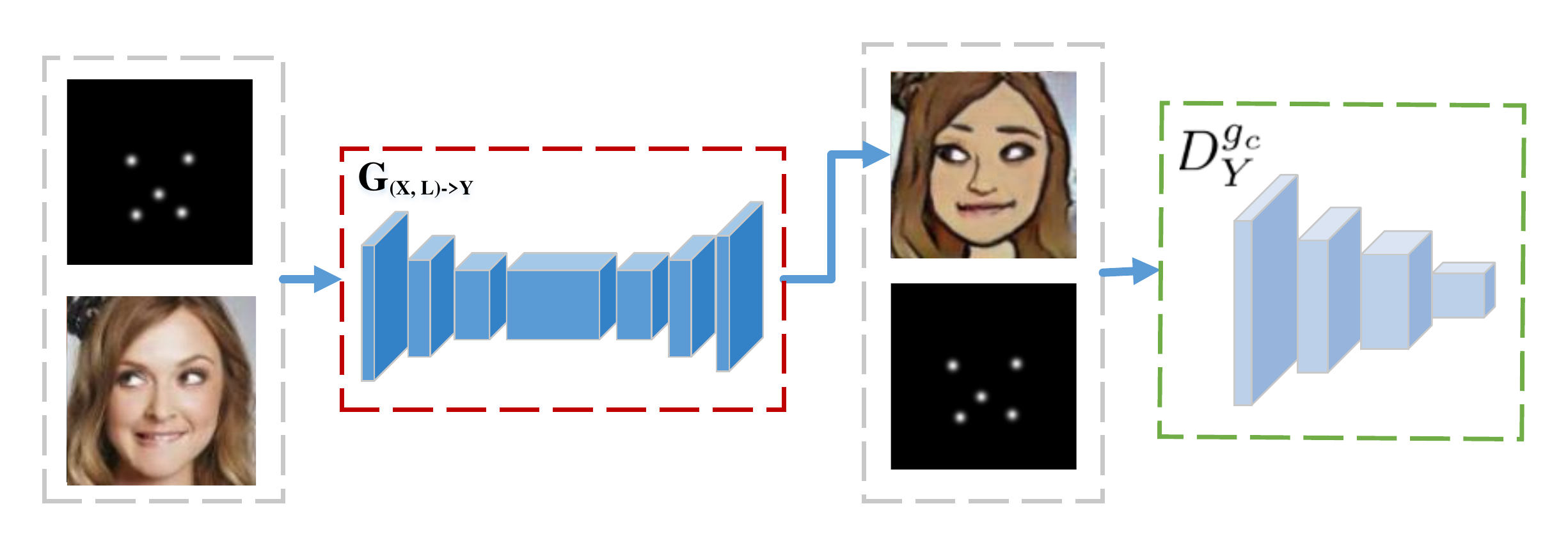}
	\caption{Global conditional discriminator. It refers to generating target images with source-domain images and landmarks as input. The generated images and its landmark predicted by the pre-trained regressor are one type of fake samples for discriminator. }
	\label{fig:landmark-match}
	\vspace{-0.12in}
\end{figure}

In order to give an explicit structure constraint between the two domains, we introduce three local discriminators on eyes, noses, and mouths respectively. The adversarial loss is defined as
\begin{equation}\label{eq:local_d}
\begin{split}
&\mathcal{L}_{GAN_{local}^{X \rightarrow Y}} = \sum_{i=1}^{3} \lambda_{l_i} \cdot 
\mathcal{L}_{GAN_{patch}}(G_{(X,L) \rightarrow Y}, D_{Y}^{l_i}) \\
&=\sum_{i=1}^{3}\lambda_{l_i} \big\{\mathbb{E}_{y} [\log D_{Y}^{l_i}(y_p)] \\
& ~~~~+ \mathbb{E}_{x}[\log (1-D_{Y}^{l_i}([{G_{(X,L) \rightarrow Y}(x)}]_p))]\big\},
\end{split}
\end{equation}
where $y_p$ and $[{G_{(X,L) \rightarrow Y}(x)}]_p$ refer to local patches of cartoon face and generated cartoon faces respectively. For the translation of $X \rightarrow Y$, the generator outputs coarse cartoon face. Then we obtain the predicted facial landmarks by the pre-trained cartoon face regressor. 

With the coordinates provided by predicted landmarks, we are able to crop local patches (eyes, nose and mouth) for local discriminators as input. Particularly, we concatenate left and right eye patches into one so that networks can learn similar sizes and colors for them. Since gradients can be back-propagated through these patches, the framework is trained in an end-to-end manner.


\subsection{Network Training}\label{sec:training}

\subsubsection{Two Stage Training}
\vspace{-0.05in}
\noindent\textbf{Stage I} First, we train our framework without local discriminators to get coarse results. At this stage, we train the generators and global discriminators involving the landmark consistency loss for two directions. This stage takes about 100K iterations where the network learns to generate a coarse result. 

\noindent\textbf{Stage II} Since we already have a coarse but reasonable result, we use the pre-trained landmark prediction network to predict facial landmarks on the coarse result. With the estimated coordinates, we extract local patches and make them input to local discriminators. Finally, we obtain a much finer result.


\vspace{-0.15in}
\subsubsection{Training setting} 
\vspace{-0.05in}
\paragraph{Cartoon Landmark Regressor Training}
We first pre-train two landmark regressor for respective domains before training the landmark assisted CycleGAN. We adopt the UNet \cite{ronneberger2015u} architecture, which takes images from different domains as input and output a 5-channel heat map as the predicted scores for facial landmarks. We train it for 80K iteration.
\vspace{-0.2in}
\paragraph{Local Patches Extraction}
We crop local patches empirically for each components, i.e. in a $128 \times 128$ image, we crop $32 \times 32$ for eye patches, $28 \times 24$ for nose patch, $23 \times 40$ for mouth patch. Thus we crop 4 patches in total (two eye patches), but the two eye patches are merged into one for discriminator. Given landmark coordinates, we extract eyes and nose patches with the corresponding landmark as center point, and extract mouth patch with two landmarks as left and right boundary. 
\vspace{-0.2in}
\paragraph{Hyper-Parameters Setting}
We set our hyper-parameters as batch size 1, initial learning rate 2e-4, and polynomial decay strategy. In the last stage, for loss hyper-parameters, $\lambda_g$ and $\lambda_{g_c}$ are both set to 0.5. $\lambda_{local}$, $\lambda_{lm}$ and $\lambda_{cyc}$ is set to 0.3, 100 and 10 respectively.

\vspace{-0.05in}
\section{Experiments}
\vspace{-0.05in}
\subsection{Dataset}
\vspace{-0.05in}
In order to accomplish our new task, we need two domains of data for cartoon and human faces. For natural human faces, we choose a classical face dataset: CelebA. For cartoon faces, we collect images of two different styles and annotate them with facial landmarks to build a new dataset.

\noindent\textbf{CelebA} For human face images, we use aligned CelebA \cite{yang2015facial} dataset for training and validation. For the whole dataset, We select the full-frontal faces, then we detect, crop and resize face images to $128\times 128$. As landmarks are provided in CelebA dataset, we do not need more operations. After our selection, we gather totally 37,794 human face images.

\noindent\textbf{Bitmoji} We collect ``bitmoji" images from Internet. We firstly annotate the landmark of the crawled images manually, i.e. the positions of eyes, mouth and nose. Then we would crop faces according to the annotated landmarks and resize them to the resolution $128\times 128$. Finally we build bitmoji style dataset of  2,125 images with its corresponding landmarks. The Bitmoji data contains rich information of human expressions and hair styles in cartoon style. With proper initialization, one can create a cartoon character which resemble the appearance of the creator.

\noindent\textbf{Anime faces} For anime faces, we follow the steps of \cite{jin2017towards} to build our dataset. First, we collect anime characters from Getchu\footnote{www.getchu.com}. Since the crawled images involve many unnecessary parts of anime, we use a pretrained cartoon face detector ``lbpcascade\_animeface"
, to detect and get bounding boxes for anime faces, then we crop the faces and resize them to size $128\times 128$. Finally, we also annotate the landmark of anime faces. Eventually we obtain a total of 17,920 images with its corresponding landmark. The anime face follows the Japanese manga style, and it is highly stylized and beautified.
\vspace{-0.1in}


\subsection{Bitmoji Faces Generation}
\vspace{-0.05in}
We firstly conduct experiments on bitmoji faces generation. Bitmoji faces are relatively similar to natural human faces in spatial structure, but there are still some obvious characteristic of bitmoji like big eyes and mouth in shape which need to be transformed in geometric structure.

As shown in Fig. \ref{bitmoji_faces_fig}, the generated results from style transfer \cite{gatys2016image} and Deep-Image-Analogy \cite{liao2017visual} are heavily affected by the selected reference images, and they temp to introduce low-level image features like texture and color of reference images to global images. MUNIT\cite{huang2018multimodal} and Improving\cite{gokaslan2018improving} get decent results as an real bitmoji image, but they temp to have the mode collapse problem and do not preserve the identity of input images, i.e. the generated results are much similar to each other. CycleGAN \cite{zhu2017unpaired} produces better results than others. Still, it does not keep sufficient attributes, loses some details and causes distortion sometimes due to insufficient consideration of geometric transformation. Our results are with higher visual quality and less visual artifacts. Besides, the generated anime faces not only contain the characteristic of ``anime" appearance, but also make the results look like the human face input to the system. 

\begin{figure}[t!]
	\centering
	\begin{tabular}
		{@{\hspace{0.0mm}}c@{\hspace{1.5mm}}c@{\hspace{1.0mm}}c@{\hspace{1.0mm}}c@{\hspace{1.0mm}}c@{\hspace{1.0mm}}c@{\hspace{1.0mm}}c@{\hspace{0mm}}}
		\rotatebox{90}{\hspace{2.5mm}\footnotesize Inputs} &
		\includegraphics[width=0.18\linewidth]{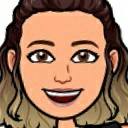}&
		\includegraphics[width=0.18\linewidth]{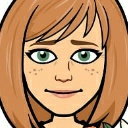}&
		\includegraphics[width=0.18\linewidth]{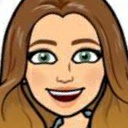}&
		\includegraphics[width=0.18\linewidth]{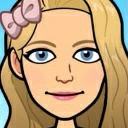}&
		\includegraphics[width=0.18\linewidth]{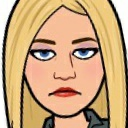}\\
		
		\rotatebox{90}{\hspace{0mm}\footnotesize CycleGAN} &
		\includegraphics[width=0.18\linewidth]{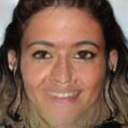}&
		\includegraphics[width=0.18\linewidth]{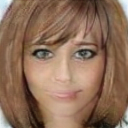}&
		\includegraphics[width=0.18\linewidth]{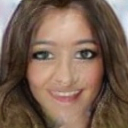}&
		\includegraphics[width=0.18\linewidth]{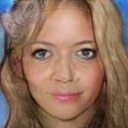}&
		\includegraphics[width=0.18\linewidth]{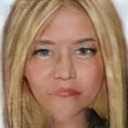}\\
		
		\rotatebox{90}{\hspace{3mm}\footnotesize Ours} &
		\includegraphics[width=0.18\linewidth]{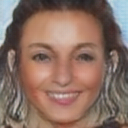}&
		\includegraphics[width=0.18\linewidth]{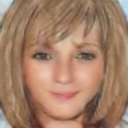}&
		\includegraphics[width=0.18\linewidth]{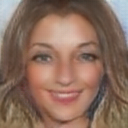}&
		\includegraphics[width=0.18\linewidth]{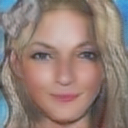}&
		\includegraphics[width=0.18\linewidth]{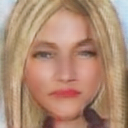}\\
		
	\end{tabular}
	\caption{Human faces translated from bitmoji faces.}
	\label{human_bitmoji_fig}
\end{figure}

In addition, translated human faces results from Bitmoji are also demonstrated in Fig. \ref{human_bitmoji_fig}. In the regions which need geometric transformation, the results generated by CycleGAN always produce blur or distortion, while our methods can get rid of this problem by introduce facial landmarks as explicit semantic correspondence.

\def\width3{0.13}
\begin{figure*}
	\centering
	\begin{tabular}
		{@{\hspace{0.0mm}}c@{\hspace{2.0mm}}c@{\hspace{1.0mm}}c@{\hspace{1.0mm}}c@{\hspace{1.0mm}}c@{\hspace{1.0mm}}c@{\hspace{1.0mm}}c@{\hspace{0mm}}}
		
		\includegraphics[width=\width3\linewidth]{figs/bitmoji/00265_inp.png} &
		\includegraphics[width=\width3\linewidth]{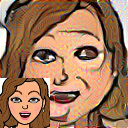} &
		\includegraphics[width=\width3\linewidth]{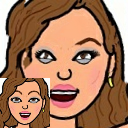} &
		\includegraphics[width=\width3\linewidth]{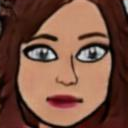}&
		\includegraphics[width=\width3\linewidth]{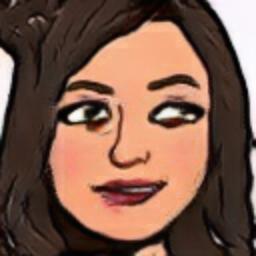}&
		\includegraphics[width=\width3\linewidth]{figs/bitmoji/00265_cyc.png}&
		\includegraphics[width=\width3\linewidth]{figs/bitmoji/00265_our.png}\\
		
		\includegraphics[width=\width3\linewidth]{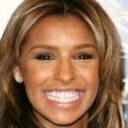} &
		\includegraphics[width=\width3\linewidth]{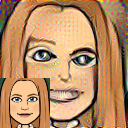} &
		\includegraphics[width=\width3\linewidth]{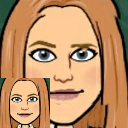} &
		\includegraphics[width=\width3\linewidth]{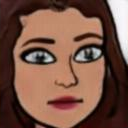}&
		\includegraphics[width=\width3\linewidth]{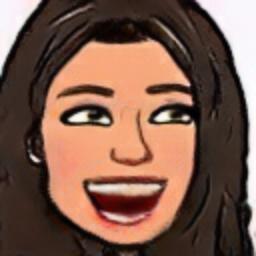}&
		\includegraphics[width=\width3\linewidth]{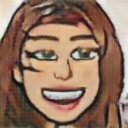}&
		\includegraphics[width=\width3\linewidth]{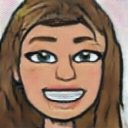}\\
		
		\includegraphics[width=\width3\linewidth]{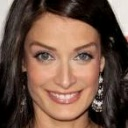} &
		\includegraphics[width=\width3\linewidth]{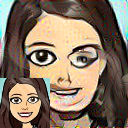} &
		\includegraphics[width=\width3\linewidth]{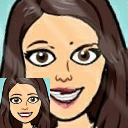} &
		\includegraphics[width=\width3\linewidth]{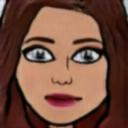}&
		\includegraphics[width=\width3\linewidth]{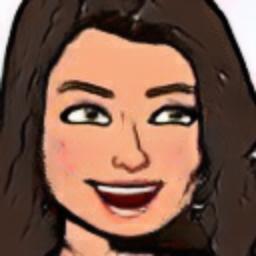}&
		\includegraphics[width=\width3\linewidth]{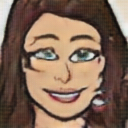}&
		\includegraphics[width=\width3\linewidth]{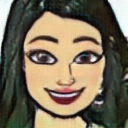}\\
		
		\includegraphics[width=\width3\linewidth]{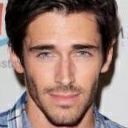} &
		\includegraphics[width=\width3\linewidth]{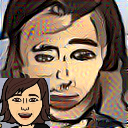} &
		\includegraphics[width=\width3\linewidth]{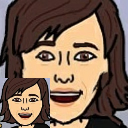} &
		\includegraphics[width=\width3\linewidth]{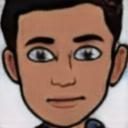}&
		\includegraphics[width=\width3\linewidth]{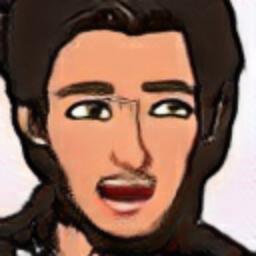}&
		\includegraphics[width=\width3\linewidth]{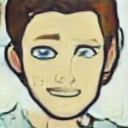}&
		\includegraphics[width=\width3\linewidth]{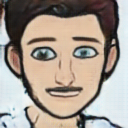}\\
		
		\includegraphics[width=\width3\linewidth]{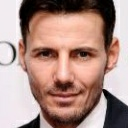} &
		\includegraphics[width=\width3\linewidth]{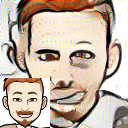} &
		\includegraphics[width=\width3\linewidth]{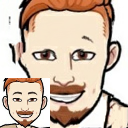} &
		\includegraphics[width=\width3\linewidth]{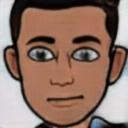}&
		\includegraphics[width=\width3\linewidth]{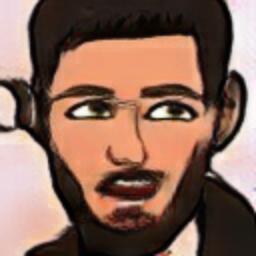}&
		\includegraphics[width=\width3\linewidth]{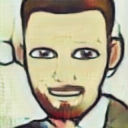}&
		\includegraphics[width=\width3\linewidth]{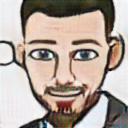}\\
		
		Inputs &
		Style \cite{gatys2016image} &
		Analogy \cite{liao2017visual}&
		Improving \cite{gokaslan2018improving}&
		MUNIT \cite{huang2018multimodal}&
		Cycle \cite{zhu2017unpaired}&
		Ours\\
		
	\end{tabular}
	\caption{Bitmoji faces generation. Style transfer\cite{gatys2016image} and Deep-Image-Analogy\cite{liao2017visual} use the reference images shown in lower left corner.}
	\label{bitmoji_faces_fig}
\end{figure*}

\subsection{Anime Faces Generation}
\vspace{-0.05in}
Translating from natural human faces to anime faces is relatively challenging since there are significant geometric changes between two domains. Thus we can see the comparison in Fig. \ref{anime_faces_fig}, without an explicit correspondence as guidance, other generation methods temp to produce a lot artifacts in the area need geometric changes, like chins, eyes, etc. 

Specifically, as shown in Fig. \ref{anime_faces_fig}, style transfer \cite{gatys2016image} and Deep-Image-Analogy \cite{liao2017visual} introduce image features like texture and color from reference image and thus the quality of them rely heavily on the reference images.  \cite{gokaslan2018improving} and MUNIT \cite{huang2018multimodal} try to get natural anime faces but temp to be trapped in mode collapse during training for this two domains.
CycleGAN \cite{zhu2017unpaired} generate decent results, but there are still artifacts in ``hard'' area which needs geometric structure changes. 

\begin{figure}[t!]
	\centering
	\begin{tabular}
		{@{\hspace{0.0mm}}c@{\hspace{1.5mm}}c@{\hspace{1.0mm}}c@{\hspace{1.0mm}}c@{\hspace{1.0mm}}c@{\hspace{1.0mm}}c@{\hspace{1.0mm}}c@{\hspace{0mm}}}
		\rotatebox{90}{\hspace{2.5mm}\footnotesize Inputs} &
		\includegraphics[width=0.18\linewidth]{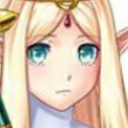}&
		\includegraphics[width=0.18\linewidth]{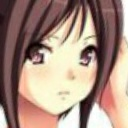}&
		\includegraphics[width=0.18\linewidth]{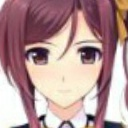}&
		\includegraphics[width=0.18\linewidth]{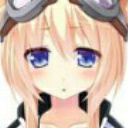}&
		\includegraphics[width=0.18\linewidth]{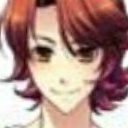}\\
		
		\rotatebox{90}{\hspace{0mm}\footnotesize CycleGAN} &
		\includegraphics[width=0.18\linewidth]{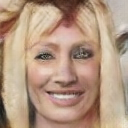}&
		\includegraphics[width=0.18\linewidth]{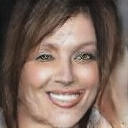}&
		\includegraphics[width=0.18\linewidth]{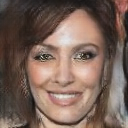}&
		\includegraphics[width=0.18\linewidth]{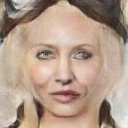}&
		\includegraphics[width=0.18\linewidth]{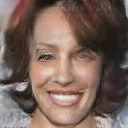}\\
		
		\rotatebox{90}{\hspace{3mm}\footnotesize Ours} &
		\includegraphics[width=0.18\linewidth]{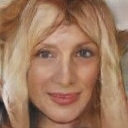}&
		\includegraphics[width=0.18\linewidth]{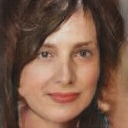}&
		\includegraphics[width=0.18\linewidth]{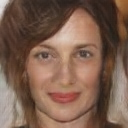}&
		\includegraphics[width=0.18\linewidth]{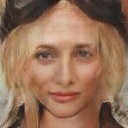}&
		\includegraphics[width=0.18\linewidth]{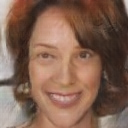}\\
		
	\end{tabular}
	\caption{Human faces translated from anime faces.}
	\label{human_anime_fig}
\end{figure}

In addition, Our results of human faces translated by anime faces are also in higher quality than CycleGAN \cite{zhu2017unpaired} and the comparisons of results are demonstrated in Fig. \ref{human_anime_fig}.

\def\width3{0.13}
\begin{figure*}
	\centering
	\begin{tabular}
		{@{\hspace{0.0mm}}c@{\hspace{2.0mm}}c@{\hspace{1.0mm}}c@{\hspace{1.0mm}}c@{\hspace{1.0mm}}c@{\hspace{1.0mm}}c@{\hspace{1.0mm}}c@{\hspace{0mm}}}

		\includegraphics[width=\width3\linewidth]{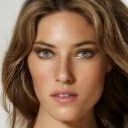} &
		\includegraphics[width=\width3\linewidth]{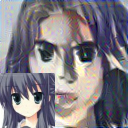} &
		\includegraphics[width=\width3\linewidth]{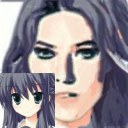} &
		\includegraphics[width=\width3\linewidth]{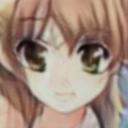}&
		\includegraphics[width=\width3\linewidth]{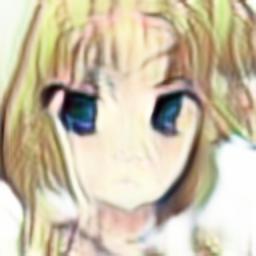}&
		\includegraphics[width=\width3\linewidth]{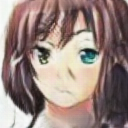}&
		\includegraphics[width=\width3\linewidth]{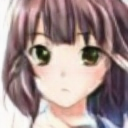}\\

		\includegraphics[width=\width3\linewidth]{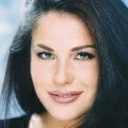} &
		\includegraphics[width=\width3\linewidth]{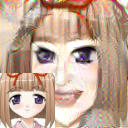} &
		\includegraphics[width=\width3\linewidth]{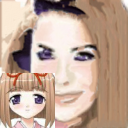} &
		\includegraphics[width=\width3\linewidth]{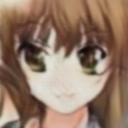}&
		\includegraphics[width=\width3\linewidth]{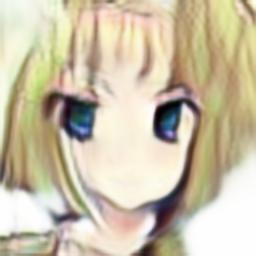}&
		\includegraphics[width=\width3\linewidth]{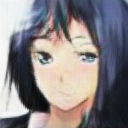}&
		\includegraphics[width=\width3\linewidth]{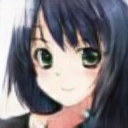}\\

		\includegraphics[width=\width3\linewidth]{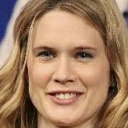} &
		\includegraphics[width=\width3\linewidth]{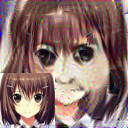} &
		\includegraphics[width=\width3\linewidth]{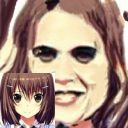} &
		\includegraphics[width=\width3\linewidth]{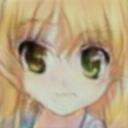}&
		\includegraphics[width=\width3\linewidth]{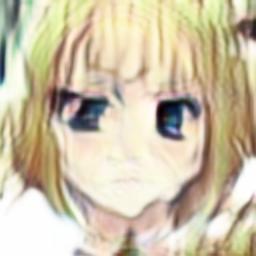}&
		\includegraphics[width=\width3\linewidth]{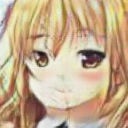}&
		\includegraphics[width=\width3\linewidth]{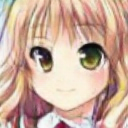}\\

		Inputs &
		Style \cite{gatys2016image} &
		Analogy \cite{liao2017visual}&
		Improving \cite{gokaslan2018improving}&
		MUNIT \cite{huang2018multimodal}&
		Cycle \cite{zhu2017unpaired}&
		Ours\\
		
	\end{tabular}
	\caption{Anime faces generation. Note that only style transfer\cite{gatys2016image} and Deep-Image-Analogy\cite{liao2017visual} use the reference images shown in lower left corner.}
	\label{anime_faces_fig}
\end{figure*}

\def\widthc{0.18}
\begin{figure}[t!]
	\centering
	\begin{tabular}
		{@{\hspace{0.0mm}}c@{\hspace{1.5mm}}c@{\hspace{1.0mm}}c@{\hspace{1.0mm}}c@{\hspace{1.0mm}}c@{\hspace{1.0mm}}c@{\hspace{1.0mm}}c@{\hspace{0mm}}}
		\rotatebox{90}{\hspace{4.0mm}\footnotesize Inputs} &
		\includegraphics[width=\widthc\linewidth]{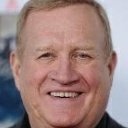}&
		\includegraphics[width=\widthc\linewidth]{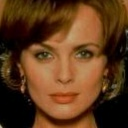}&
		\includegraphics[width=\widthc\linewidth]{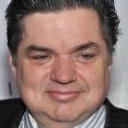}&
		\includegraphics[width=\widthc\linewidth]{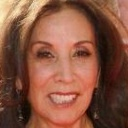}&
		\includegraphics[width=\widthc\linewidth]{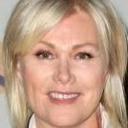}\\

		\rotatebox{90}{\hspace{0.2mm}\footnotesize CycleGAN} &
		\includegraphics[width=\widthc\linewidth]{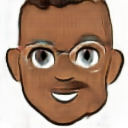}&
		\includegraphics[width=\widthc\linewidth]{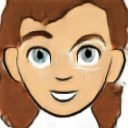}&
		\includegraphics[width=\widthc\linewidth]{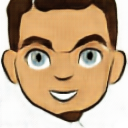}&
		\includegraphics[width=\widthc\linewidth]{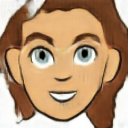}&
		\includegraphics[width=\widthc\linewidth]{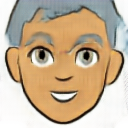}\\

		\rotatebox{90}{\hspace{4.8mm}\footnotesize Ours} &
		\includegraphics[width=\widthc\linewidth]{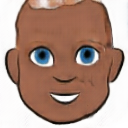}&
		\includegraphics[width=\widthc\linewidth]{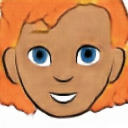}&
		\includegraphics[width=\widthc\linewidth]{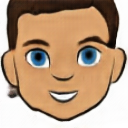}&
		\includegraphics[width=\widthc\linewidth]{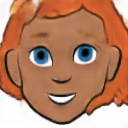}&
		\includegraphics[width=\widthc\linewidth]{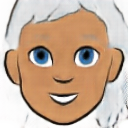}\\

	\end{tabular}
	\caption{Our failure cases on Cartoonset10K~\cite{cart10k} faces generation. It indicates the limitation of our framework on such dataset with few variation.}
	\label{cartoon10k_fig}
	\vspace{-0.2in}
\end{figure}
\vspace{-0.05in}
\subsection{Ablation Study}
\vspace{-0.05in}
We conduct ablation study for our framework and take the translation from human faces to anime faces as our examples since geometric variation are more significant between these two domains. We mainly study the effectiveness of landmark, local discriminator and training strategy separately. In the experiments, CycleGAN is the `Basic Model' of our framework.

\vspace{-0.15in}
\subsubsection{Landmark Conditions and Landmark Consistency Loss}
\vspace{-0.05in}
Following the human-pose estimation task, we encode the coordinates into heat map consisting of a 2D Gaussian centered at the key-point location. In order to predict landmark of generated images, we first train the landmark prediction network. To make a comparison with basic model and verify the effect of landmark embedding condition as well as landmark consistency loss, we first feed landmark heat maps to both generator and conditional discriminator as part of input. Results are shown in the third row \emph{Lm\_cd} in Fig. \ref{comb_fig}, where the facial structures (eyes, nose and mouth) are clearer than the basic model.

Then landmark consistency loss is added to make constraints on the facial landmark between input and its translated results. We add such consistency constraint to our basic model and the results are shown in the fourth row \emph{Lm\_co} in Fig. \ref{comb_fig}. Compared to \emph{Lm\_cd}, \emph{Lm\_co} can keep good results for facial structures and reduce visual artifacts compared to Basic model. All the results are shown in Fig. \ref{comb_fig}.

\vspace{-0.15in}
\subsubsection{Local Discriminator}
\vspace{-0.1in}
Local discriminators contain eye-patch discriminator, nose-patch discriminator and mouth-patch discriminator. We then discard local discriminators from the whole framework to verify the effects of local discriminators. Results are shown in Fig. \ref{comb_fig} where \emph{w/o-local} refers to frameworks without local discriminators while \emph{Full} refers to our whole framework. The complete framework with local discriminator generates high-quality images with less problems and more details on facial structures around eyes, nose and mouth.

\vspace{-0.15in}
\subsubsection{Pretraining Analysis}
\vspace{-0.05in}
\paragraph{Two Stage Training}
We train our model with two stages to make a more stable framework. To analyse the role of two stage training, we conduct experiment with only one whole stage instead of two-stage training. The results are shown in Fig. \ref{pretrain_fig} (2nd col). Although results looks decent but are clearly inferior to results of two-stage training.

\vspace{-0.15in}
\paragraph{Landmark Prediction Network Pretraining}
In our framework, landmark prediction network is pretrained in an inital phase. Since local patches cannot be extracted correctly and landmark consistency loss will contribute nothing to the translation network since the predicted landmarks are generated randomly without a resonable pretraining of landmark prediction network at the beginning of training. To verify the role of landmark prediction network pre-training, we conduct experiment to train landmark prediction network and translation network simultaneously without pre-training and the results are shown in Fig. \ref{pretrain_fig} (3 col). 

	\def\width3{0.12}
	\begin{figure}[h]
		\centering
		\begin{tabular}
			{@{\hspace{0.0mm}}c@{\hspace{1.0mm}}c@{\hspace{0.0mm}}c@{\hspace{0.0mm}}c@{\hspace{2.0mm}}c@{\hspace{1.0mm}}c@{\hspace{0.0mm}}c@{\hspace{0.0mm}}c@{\hspace{0mm}}}
			
			\includegraphics[width=\width3\linewidth]{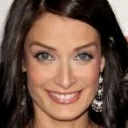} &
			\includegraphics[width=\width3\linewidth]{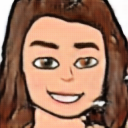} &
			\includegraphics[width=\width3\linewidth]{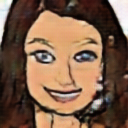} &
			\includegraphics[width=\width3\linewidth]{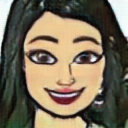}&
			
			\includegraphics[width=\width3\linewidth]{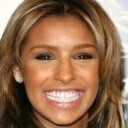} &
			\includegraphics[width=\width3\linewidth]{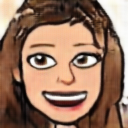} &
			\includegraphics[width=\width3\linewidth]{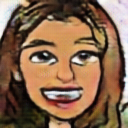} &
			\includegraphics[width=\width3\linewidth]{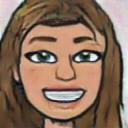}\\
			
			\includegraphics[width=\width3\linewidth]{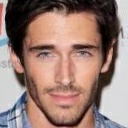} &
			\includegraphics[width=\width3\linewidth]{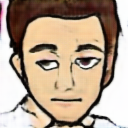} &
			\includegraphics[width=\width3\linewidth]{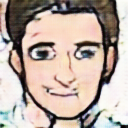} &
			\includegraphics[width=\width3\linewidth]{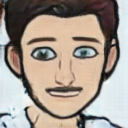}&
			
			\includegraphics[width=\width3\linewidth]{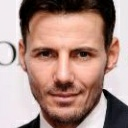} &
			\includegraphics[width=\width3\linewidth]{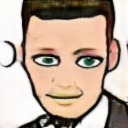} &
			\includegraphics[width=\width3\linewidth]{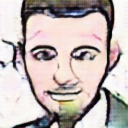} &
			\includegraphics[width=\width3\linewidth]{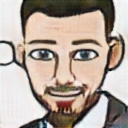}\\

		\end{tabular}
		\caption{Ablation study on the role of stage 1 coarse training and pretraining for landmark prediction network. For each example, we show input (1st col.), results of training with one whole stage (2nd col.), results without pre-training of landmark prediction network (3rd col.) and our results (last col.).}
		\label{pretrain_fig}
	\end{figure}

\def\width{0.18}

\begin{figure}[!t]
	\centering
	\begin{tabular}
		{@{\hspace{0.0mm}}c@{\hspace{2mm}}c@{\hspace{1.2mm}}c@{\hspace{1.2mm}}c@{\hspace{1.2mm}}c@{\hspace{1.2mm}}c@{\hspace{0mm}}}
		\rotatebox{90}{\hspace{2.5mm}\footnotesize Inputs} &
		\includegraphics[width=\width\linewidth]{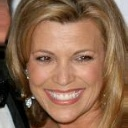}&
		\includegraphics[width=\width\linewidth]{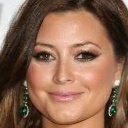}&
		\includegraphics[width=\width\linewidth]{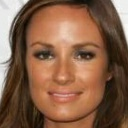}&
		\includegraphics[width=\width\linewidth]{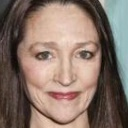}&
		\includegraphics[width=\width\linewidth]{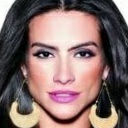}\\
		
		\rotatebox{90}{\hspace{3mm}\footnotesize Basic} &
		\includegraphics[width=\width\linewidth]{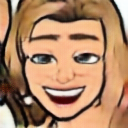}&
		\includegraphics[width=\width\linewidth]{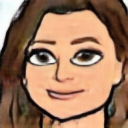}&
		\includegraphics[width=\width\linewidth]{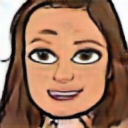}&
		\includegraphics[width=\width\linewidth]{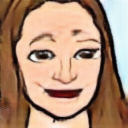}&
		\includegraphics[width=\width\linewidth]{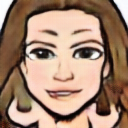}\\
		
		\rotatebox{90}{\hspace{2.5mm}\footnotesize Lm\_cd} &
		\includegraphics[width=\width\linewidth]{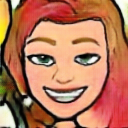}&
		\includegraphics[width=\width\linewidth]{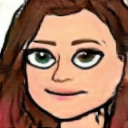}&
		\includegraphics[width=\width\linewidth]{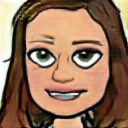}&
		\includegraphics[width=\width\linewidth]{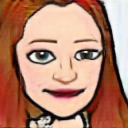}&
		\includegraphics[width=\width\linewidth]{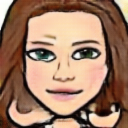}\\
		
		\rotatebox{90}{\hspace{2.5mm}\footnotesize Lm\_co} &
		\includegraphics[width=\width\linewidth]{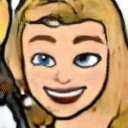}&
		\includegraphics[width=\width\linewidth]{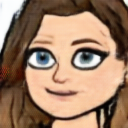}&
		\includegraphics[width=\width\linewidth]{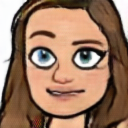}&
		\includegraphics[width=\width\linewidth]{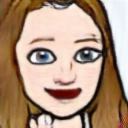}&
		\includegraphics[width=\width\linewidth]{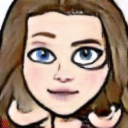}\\
		
		\rotatebox{90}{\hspace{1mm}\footnotesize w/o-local} &
		\includegraphics[width=\width\linewidth]{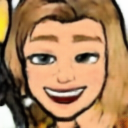}&
		\includegraphics[width=\width\linewidth]{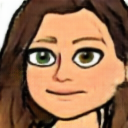}&
		\includegraphics[width=\width\linewidth]{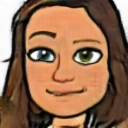}&
		\includegraphics[width=\width\linewidth]{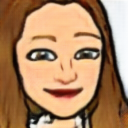}&
		\includegraphics[width=\width\linewidth]{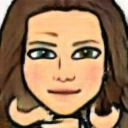}\\
		
		\rotatebox{90}{\hspace{3.5mm} \footnotesize Full} &
		\includegraphics[width=\width\linewidth]{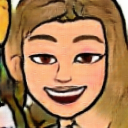}&
		\includegraphics[width=\width\linewidth]{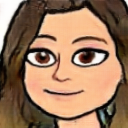}&
		\includegraphics[width=\width\linewidth]{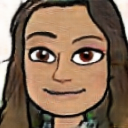}&
		\includegraphics[width=\width\linewidth]{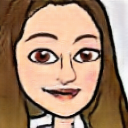}&
		\includegraphics[width=\width\linewidth]{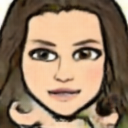}\\
		
	\end{tabular}\\
	\caption{Ablation study on landmark assisted parts.}
	\label{comb_fig}
	\vspace{-0.15in}
\end{figure}

\begin{table*}[t!]
	\centering
	\caption{Results of user study on generated bitmoji faces samples.}
	\label{user}
	\begin{tabular}
		{c|c|c|c|c|c|c|c}
		\hline
		~~~~~~& Rank&
		Style \cite{gatys2016image} &
		Analogy \cite{liao2017visual} &
		Improving \cite{gokaslan2018improving} &
		MUNIT \cite{huang2018multimodal}&
		CycleGAN \cite{zhu2017unpaired} &
		Ours\\
		\hline

		\begin{tabular}
			{c}
			Identity
		\end{tabular}
		&
		\begin{tabular}
			{c}
			Top1 \\
			
			Top3 \\
		\end{tabular}
		&
		\begin{tabular}
			{c}
			0.00 \\
			
			~~0.00~~ \\
		\end{tabular}
		&
		\begin{tabular}
			{c}
			0.26 \\
			
			~~~0.14~~~ \\
		\end{tabular}
		&
		\begin{tabular}
			{c}
			0.00 \\
			
			~~0.06~~ \\
		\end{tabular}
		&
		\begin{tabular}
			{c}
			0.00 \\
			
			~~0.20~~ \\
		\end{tabular}
		&
		\begin{tabular}
			{c}
			0.00 \\
			
			~~0.14~~ \\
		\end{tabular}
		&
		\begin{tabular}
			{c}
			\textbf{0.74} \\
			
			~~\textbf{0.46}~~ \\
		\end{tabular} \\
		\hline
		
		\begin{tabular}
			{c}
			Realistic
		\end{tabular}
		&
		\begin{tabular}
			{c}
			Top1 \\
			
			Top3 \\
		\end{tabular}
		&
		\begin{tabular}
			{c}
			0.00 \\
			
			~~0.00~~~ \\
		\end{tabular}
		&
		\begin{tabular}
			{c}
			0.00 \\
			
			~~~0.22~~~ \\
		\end{tabular}
		&
		\begin{tabular}
			{c}
			0.48 \\
			
			~~\textbf{0.30}~~ \\
		\end{tabular}
		&
		\begin{tabular}
			{c}
			0.00 \\
			
			~~0.17~~ \\
		\end{tabular}
		&
		\begin{tabular}
			{c}
			0.00 \\
			
			~~0.00~~ \\
		\end{tabular}
		&
		\begin{tabular}
			{c}
			\textbf{0.52} \\
			
			~~\textbf{0.30}~~ \\
		\end{tabular} \\
		\hline
		\begin{tabular}
			{c}
			AsProfile
		\end{tabular}
		&
		\begin{tabular}
			{c}
			Top1 \\
			
			Top3 \\
		\end{tabular}
		&
		\begin{tabular}
			{c}
			0.00 \\
			
			~~0.00~~~ \\
		\end{tabular}
		&
		\begin{tabular}
			{c}
			0.00 \\
			
			~~~0.05~~~ \\
		\end{tabular}
		&
		\begin{tabular}
			{c}
			0.00 \\
			
			~~0.20~~ \\
		\end{tabular}
		&
		\begin{tabular}
			{c}
			0.23 \\
			
			~~0.26~~ \\
		\end{tabular}
		&
		\begin{tabular}
			{c}
			0.00 \\
			
			~~0.06~~ \\
		\end{tabular}
		&
		\begin{tabular}
			{c}
			\textbf{0.77} \\
			
			~~\textbf{0.42}~~ \\
		\end{tabular} \\
		\hline
	\end{tabular}
\end{table*}

\begin{table}[t!]
	\centering
	\caption{Quantitative comparisons on different methods and components.}
	\label{quantiative}
	\begin{tabular}
		{c|c}
		\hline
		Methods & FID \\
		\hline
		Style \cite{gatys2016image} & 13509.25 \\
		Analogy \cite{liao2017visual} & 11933.63 \\
		Improving \cite{gokaslan2018improving} & 10365.39 \\
		MUNIT \cite{huang2018multimodal} & 2749.46 \\
		CycleGAN \cite{zhu2017unpaired} & 2398.16 \\
		\hline
		Ours\_Lm\_cd & 2140.88  \\
		Ours\_Lm\_co & 2286.39 \\
		Ours\_w/o-local & 1993.83 \\
		Ours\_Full & \textbf{1988.50} \\
		\hline
	\end{tabular}
	\vspace{-0.1in}
\end{table}

\vspace{-0.1in}
\subsection{Quantitative Study}
\vspace{-0.05in}
Although we gain high visual quality among generated images, we also utilize quantitative metrics to evaluate our results. To evaluate the difference between our generated image and an anime face, we adopt the Fréchet Inception Distance (FID) \cite{heusel2017gans}. Following the steps in \cite{jin2017towards}, we calculate a 4096D feature vector by a pre-trained network \cite{saito2015illustration2vec} for each test anime face. The whole test set contains totally 2,000 samples. We calculate the mean and covariance matrices of the 4096D feature vector for results generated by different methods and anime face test set respectively. Then we calculate FID for different methods. The comparison is shown in Tab. \ref{quantiative}.

The results in Tab. \ref{quantiative} reveal that our methods get the minimum FID, which means our generated results have a closest distribution with real anime faces and thus the results of our images are like real anime faces most. 

\subsection{User Study}
To further prove the effectiveness of methods, we conduct a user study for method comparison. We mainly consider three aspects for the generated results: \emph{Identity}, indicates whether our generated results keeps the identity of input human faces. \emph{Realistic}, refers to whether the generated results looks real in target domain. i.e. looks like anime faces. \emph{AsProfile}, an overall evaluation from user, indicates whether it is a good result as their profile. 

We set 48 groups of bitmoji samples and conduct user study among 59 users, who are required to pick 3 sorted results from the methods. The top-1 and top-3 rates are shown in Tab. \ref{user}. From the Tab. \ref{user} we can notice that our methods all rank first in three metrics, which means our results not only looks like real bitmoji faces but also preserve identity of inputs. In addition to our method, Analogy \cite{liao2017visual} can keep identities well, Improving \cite{gokaslan2018improving} looks more like cartoon faces except our methods. The generated results by our method and MUNIT \cite{huang2018multimodal} may be the most popular among users.

\subsection{Discussion}
With the experiments on Bitmoji and anime faces generation, we found that the characteristic of a dataset plays an important role to the translated image quality.

In general, GANs-like methods such as \cite{zhu2017unpaired, gokaslan2018improving, huang2018multimodal} would require sample images from two domains aligned well to generate good results, thus in Bitmoji generation experiments, they can generate decent translated results in aligned region but fail in regions with mismatched geometric structure. Instead, with the extra landmarks as constraints and conditions, ours eliminate these artifacts and preserve the structures. Similarly, our method also largely outperforms GAN-like schemes in anime face generation though larger geometric inconsistency existed.

In addition, the limited variance among samples in dataset can affect the identity preserving of generated results. We take the cartoonset10k dataset for the cartoon generation. As the result shown in Fig. \ref{cartoon10k_fig}, all the translated images are similar in appearance with different inputs. This is because the training samples of this dataset are generated only by combinations of some fixed components which is in-sufficient for representing the characteristics of natural human faces. This is a limitation of our framework to be addressed in the near future.

\section{Conclusion}
In this paper, we have proposed a method to generate cartoon faces based on input human faces by utilizing unpaired training data. Since there are huge geometric and structural differences between these two types of face images, we introduced landmark assisted CycleGAN, which utilizes facial landmarks to constrain the facial structure between two domains and guide the training of local discriminators. Since cartoon faces and its corresponding landmarks are not accessible from public data, we build a dataset involving 17,920 samples for anime faces style and 2,125 samples for bitmoji style. Finally, by training our network, impressive high quality cartoon faces and bitmojis are generated.

For now, we only generate high quality images with relatively low resolution. It would be a challenging task to make them high-resolution as well as detail-rich. But we will take it as part of our future work. 

\clearpage
{\small
\bibliographystyle{ieee}
\bibliography{egbib}
}
\clearpage

\section{Appendix}
\subsection{Details of Different Network Architecture}
\subsubsection{Landmark Regressor Network}
Landmark regressor network is an U-Net \cite{ronneberger2015u} like network for predicting facial landmarks in both source and target domains. Firstly, it will be pre-trained in cartoon faces and real faces domain respectively. 
Secondly, during training of the whole framework, the network is utilized to predict landmark of results generated by generators. 
The network architecture is shown in Tab. \ref{pred}.

\begin{table}[h]
	\centering
	\scalebox{0.88}{
		\begin{tabular}
			{|c|c|c|}
			\hline
			Layer & Output Size & ~(Kernel, Stride)~\\
			\hline
			Inputs & $128\times128\times3$ & (- , -) \\
			\hline
			Conv1 & $64\times64\times64$ & (3, 2) \\
			\hline
			Conv2 & $32\times32\times128$ & (3, 2) \\
			\hline
			Conv3 & $16\times16\times256$ & (3, 2) \\
			\hline
			Conv4 & $8\times8\times512$ & (3, 2) \\
			\hline
			Conv5 & $4\times4\times1024$ & (3, 2) \\
			\hline
			Resblock1 & $4\times4\times1024$ & (3, 1) \\
			\hline
			~~Concat(Deconv5, Conv4)~~ & $8\times8\times512$ & (3, 2) \\
			\hline
			~~Concat(Deconv4, Conv3)~~ & $16\times16\times256$ & (3, 2)\\
			\hline
			~~Concat(Deconv3, Conv2)~~ & $32\times32\times128$ & (3, 2)\\
			\hline
			~~Concat(Deconv2, Conv1)~~ & $64\times64\times64$ & (3, 2) \\
			\hline
			~~Concat(Deconv1, Inputs)~~ & $128\times128\times32$ & (3, 2) \\
			\hline
			Conv\_output1 & $128\times128\times32$ & (3, 1) \\
			\hline
			Conv\_output2 & $128\times128\times3$ & (3, 1) \\
			\hline
	\end{tabular}}
	\vspace{0.1in}
	\caption{Network architecture of landmark regressor network, where `Concat' means two feature maps are concatenated along the channel axis, `Conv$i$' and `Deconv$i$' refer to a convolution and a deconvolution layer respectively.}
	\label{pred}
\end{table}

\subsubsection{Conditional Global Discriminator}
Since we generate target images with source-domain images and corresponding landmarks as input, thus for conditional global discriminator, only target images with corresponding correct landmarks are viewed as real samples, while the samples with generated images or unmatched landmarks are viewed as fake samples. The architecture of conditional global discriminator is shown in Tab. \ref{global_d_tab}.

\begin{table}[h]
	\centering
	\begin{tabular}
		{|c|c|c|}
		\hline
		Layer & Output Size & (Kernel, Stride)\\
		\hline
		Inputs & $128\times128\times3$ & (- , -)\\
		\hline
		Conv1 & $64\times64\times64$ & (4, 2) \\
		\hline
		Conv2 & $32\times32\times128$ & (4, 2) \\
		\hline
		Conv3 & $16\times16\times256$ & (4, 2) \\
		\hline
		Conv4 & $8\times8\times512$ & (4, 2) \\
		\hline
		Fc1 & $1$ & (1,1)\\
		\hline
	\end{tabular}
	\vspace{0.1in}
	\caption{Network architecture of the conditional global discriminator.}
	\label{global_d_tab}
\end{table}

\subsection{More Results for Bitmoji Faces Generation}
\paragraph{Comparison with \cite{wolf2017unsupervised}}
Since Wolf et al. \cite{wolf2017unsupervised} achieve state-of-the-art results on bitmoji faces generation, to make a comparison with their method is necessary. 
However, their codes are not open-source, and to re-implement the method needs additional training data which is unavailable. 
We thus directly crop the input images from original paper of \cite{wolf2017unsupervised} and then apply our method to their inputs. The comparisons are shown in Fig. \ref{wolf}.
Although this comparison may not be entirely fair, we can see that our results preserve better facial geometry and expressions than the results of Wolf et al. \cite{wolf2017unsupervised}.

\def\width3{0.16}
\begin{figure}[h]
	\centering
	\begin{tabular}
		{@{\hspace{0.0mm}}c@{\hspace{1.0mm}}c@{\hspace{0.0mm}}c@{\hspace{2.0mm}}c@{\hspace{1.0mm}}c@{\hspace{0.0mm}}c@{\hspace{0mm}}}

		\includegraphics[width=\width3\linewidth]{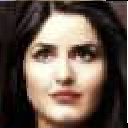} &
		\includegraphics[width=\width3\linewidth]{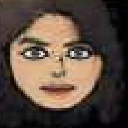} &
		\includegraphics[width=\width3\linewidth]{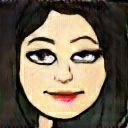} &

		\includegraphics[width=\width3\linewidth]{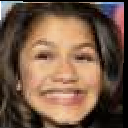} &
		\includegraphics[width=\width3\linewidth]{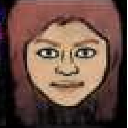} &
		\includegraphics[width=\width3\linewidth]{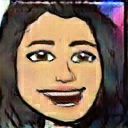} \\
		
		\includegraphics[width=\width3\linewidth]{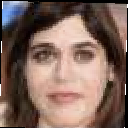} &
		\includegraphics[width=\width3\linewidth]{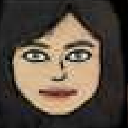} &
		\includegraphics[width=\width3\linewidth]{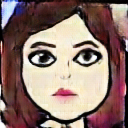} &
		
		\includegraphics[width=\width3\linewidth]{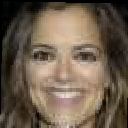} &
		\includegraphics[width=\width3\linewidth]{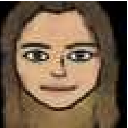} &
		\includegraphics[width=\width3\linewidth]{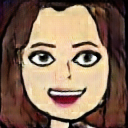} \\

	\end{tabular}
	\vspace{0.1in}
	\caption{Comparison with \cite{wolf2017unsupervised}. For each example, we show the inputs (1st col.), results from Wolf et al. \cite{wolf2017unsupervised} (2nd col.), and our results (last col.).}
	\label{wolf}
\end{figure}

\paragraph{More Visual Results}
More visual results for bitmoji faces generation are shown in Fig.~\ref{bitmoji} and we also compare some of our results with CycleGAN~\cite{zhu2017unpaired} in Fig.~\ref{bitmoji2}.

\subsection{More Results for Anime Faces Generation}
More visual results for anime faces generation are shown in Fig.~\ref{anime} and we make a comparison with CycleGAN~\cite{zhu2017unpaired} on examples in Fig.~\ref{anime2}.

\def\widtht{0.29}
\begin{figure*}[t!]
	\centering
	\begin{tabular}
		{@{\hspace{0.0mm}}c@{\hspace{2.5mm}}c@{\hspace{2.5mm}}c@{\hspace{0mm}}}
		
		\includegraphics[width=\widtht\linewidth]{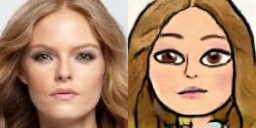} &
		\includegraphics[width=\widtht\linewidth]{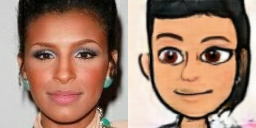} &
		\includegraphics[width=\widtht\linewidth]{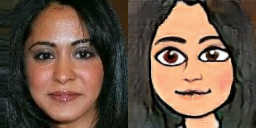} \\
		
		\includegraphics[width=\widtht\linewidth]{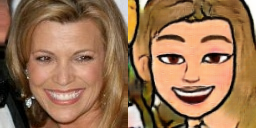} &
		\includegraphics[width=\widtht\linewidth]{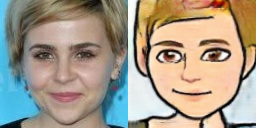} &
		\includegraphics[width=\widtht\linewidth]{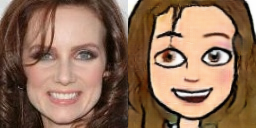} \\
		
		\includegraphics[width=\widtht\linewidth]{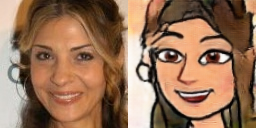} &
		\includegraphics[width=\widtht\linewidth]{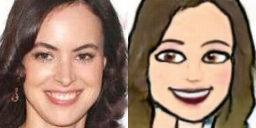} &
		\includegraphics[width=\widtht\linewidth]{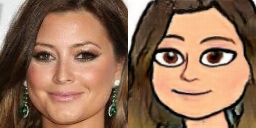} \\
		
		\includegraphics[width=\widtht\linewidth]{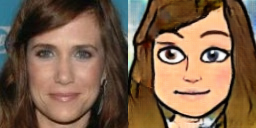} &
		\includegraphics[width=\widtht\linewidth]{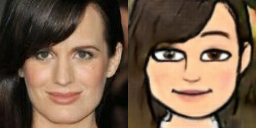} &
		\includegraphics[width=\widtht\linewidth]{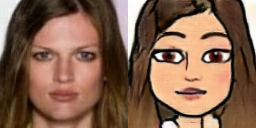} \\
		
		\includegraphics[width=\widtht\linewidth]{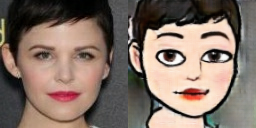} &
		\includegraphics[width=\widtht\linewidth]{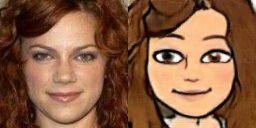} &
		\includegraphics[width=\widtht\linewidth]{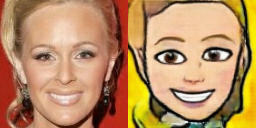} \\
		
		\includegraphics[width=\widtht\linewidth]{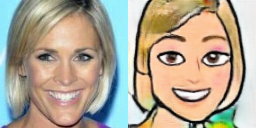} &
		\includegraphics[width=\widtht\linewidth]{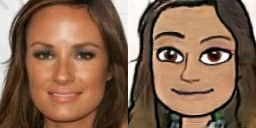} &
		\includegraphics[width=\widtht\linewidth]{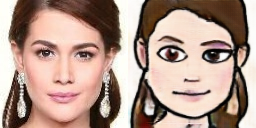} \\
		
		\includegraphics[width=\widtht\linewidth]{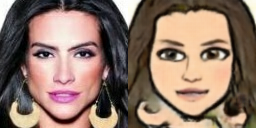} &
		\includegraphics[width=\widtht\linewidth]{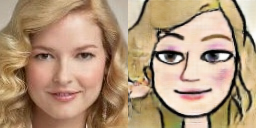} &
		\includegraphics[width=\widtht\linewidth]{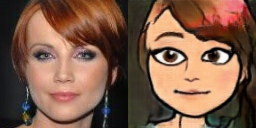} \\
		
		\includegraphics[width=\widtht\linewidth]{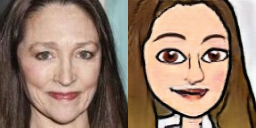} &
		\includegraphics[width=\widtht\linewidth]{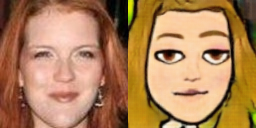} &
		\includegraphics[width=\widtht\linewidth]{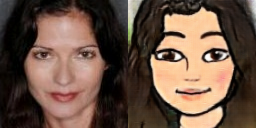} \\
		
		\begin{tabular}{cc} Inputs~~ & ~~~~~~~~Results \end{tabular} &
		\begin{tabular}{cc} Inputs~~ & ~~~~~~~~Results \end{tabular} &
		\begin{tabular}{cc} Inputs~~ & ~~~~~~~~Results \end{tabular} \\

	\end{tabular}
	\vspace{0.1in}
	\caption{More Bitmoji faces generation.}
	\label{bitmoji}
\end{figure*}

\begin{figure*}[t!]
	\centering
	\begin{tabular}
		{@{\hspace{0.0mm}}c@{\hspace{2.5mm}}c@{\hspace{2.5mm}}c@{\hspace{0mm}}}
		
		\includegraphics[width=\widtht\linewidth]{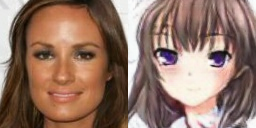} &
		\includegraphics[width=\widtht\linewidth]{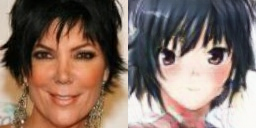} &
		\includegraphics[width=\widtht\linewidth]{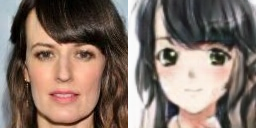} \\
		
		\includegraphics[width=\widtht\linewidth]{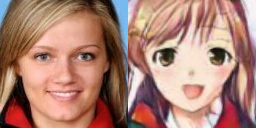} &
		\includegraphics[width=\widtht\linewidth]{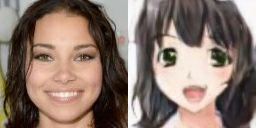} &
		\includegraphics[width=\widtht\linewidth]{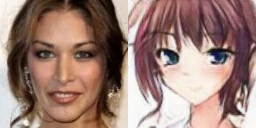} \\
		
		\includegraphics[width=\widtht\linewidth]{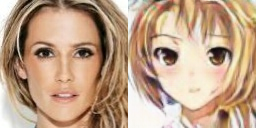} &
		\includegraphics[width=\widtht\linewidth]{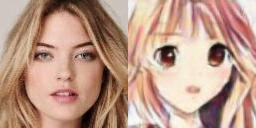} &
		\includegraphics[width=\widtht\linewidth]{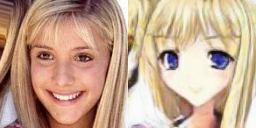} \\
		
		\includegraphics[width=\widtht\linewidth]{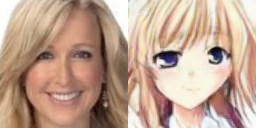} &
		\includegraphics[width=\widtht\linewidth]{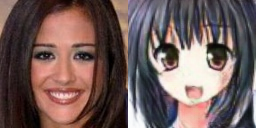} &
		\includegraphics[width=\widtht\linewidth]{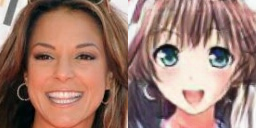} \\
		
		\includegraphics[width=\widtht\linewidth]{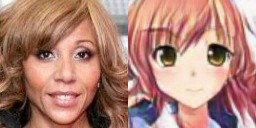} &
		\includegraphics[width=\widtht\linewidth]{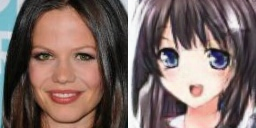} &
		\includegraphics[width=\widtht\linewidth]{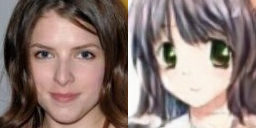} \\
		
		\includegraphics[width=\widtht\linewidth]{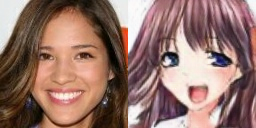} &
		\includegraphics[width=\widtht\linewidth]{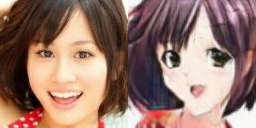} &
		\includegraphics[width=\widtht\linewidth]{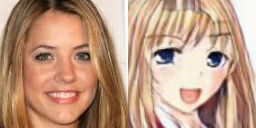} \\
		
		\includegraphics[width=\widtht\linewidth]{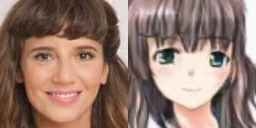} &
		\includegraphics[width=\widtht\linewidth]{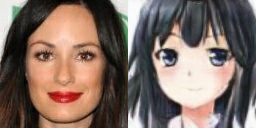} &
		\includegraphics[width=\widtht\linewidth]{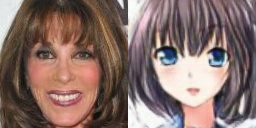} \\
		
		\includegraphics[width=\widtht\linewidth]{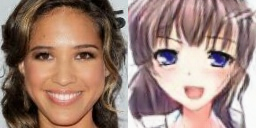} &
		\includegraphics[width=\widtht\linewidth]{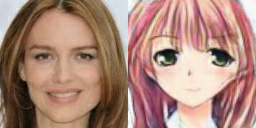} &
		\includegraphics[width=\widtht\linewidth]{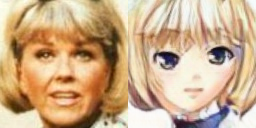} \\
		
		\begin{tabular}{cc} Inputs~~ & ~~~~~~~~Results \end{tabular} &
		\begin{tabular}{cc} Inputs~~ & ~~~~~~~~Results \end{tabular} &
		\begin{tabular}{cc} Inputs~~ & ~~~~~~~~Results \end{tabular} \\

	\end{tabular}
	\vspace{0.1in}
	\caption{More Anime faces generation.}
	\label{anime}
\end{figure*}

\def\widtht{0.45}
\begin{figure*}[t!]
	\centering
	\begin{tabular}
		{@{\hspace{0.0mm}}c@{\hspace{2.5mm}}c@{\hspace{0mm}}}
		
		\includegraphics[width=\widtht\linewidth]{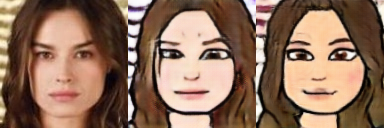} &
		\includegraphics[width=\widtht\linewidth]{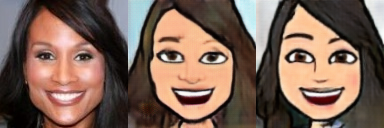} \\

		\includegraphics[width=\widtht\linewidth]{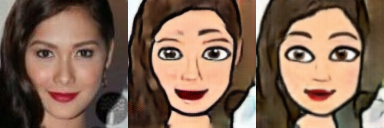} &
		\includegraphics[width=\widtht\linewidth]{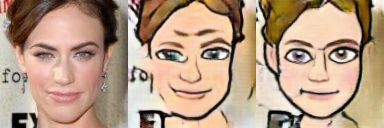} \\
		
		\includegraphics[width=\widtht\linewidth]{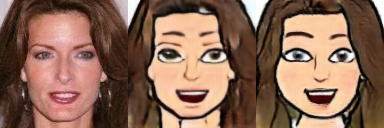} &
		\includegraphics[width=\widtht\linewidth]{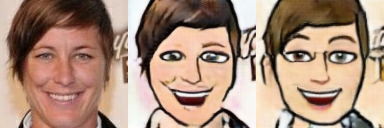} \\
		
		\includegraphics[width=\widtht\linewidth]{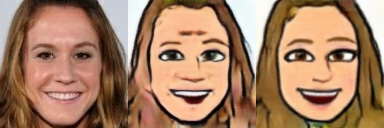} &
		\includegraphics[width=\widtht\linewidth]{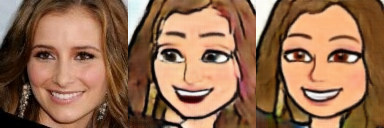} \\

		\begin{tabular}{ccc} Inputs~~~~~~~ & ~~~CycleGAN\cite{zhu2017unpaired}~~~ & ~~~~~~~Ours \end{tabular} &
		\begin{tabular}{ccc} Inputs~~~~~~~ & ~~~CycleGAN\cite{zhu2017unpaired}~~~ & ~~~~~~~Ours \end{tabular} \\

	\end{tabular}
	\vspace{0.1in}
	\caption{More Bitmoji faces generation compared with CycleGAN \cite{zhu2017unpaired}.}
	\label{bitmoji2}
\end{figure*}
\vspace{0.3in}

\def\widtht{0.45}
\begin{figure*}[t!]
	\centering
	\begin{tabular}
		{@{\hspace{0.0mm}}c@{\hspace{2.5mm}}c@{\hspace{0mm}}}
		
		\includegraphics[width=\widtht\linewidth]{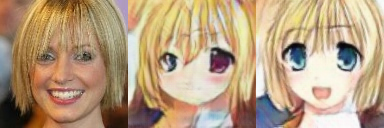} &
		\includegraphics[width=\widtht\linewidth]{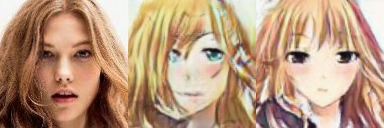} \\
		
		\includegraphics[width=\widtht\linewidth]{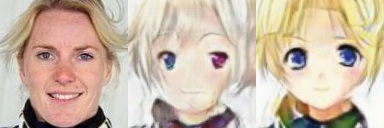} &
		\includegraphics[width=\widtht\linewidth]{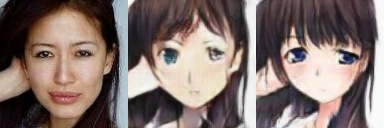} \\
		
		\includegraphics[width=\widtht\linewidth]{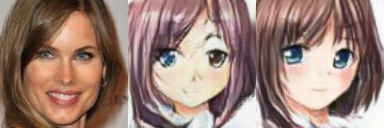} &
		\includegraphics[width=\widtht\linewidth]{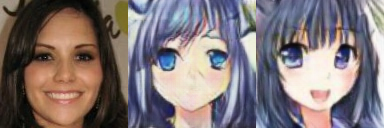} \\

		\begin{tabular}{ccc} Inputs~~~~~~~ & ~~~CycleGAN\cite{zhu2017unpaired}~~~ & ~~~~~~~Ours \end{tabular} &
		\begin{tabular}{ccc} Inputs~~~~~~~ & ~~~CycleGAN\cite{zhu2017unpaired}~~~ & ~~~~~~~Ours \end{tabular} \\

	\end{tabular}
	\vspace{0.1in}
	\caption{More Anime faces generation compared with CycleGAN \cite{zhu2017unpaired}.}
	\label{anime2}
\end{figure*}

\end{document}